\documentclass{ieeeaccess}
\usepackage{cite}
\usepackage{amsmath,amssymb,amsfonts}
\usepackage{algorithm}
\usepackage{algpseudocode}
\usepackage{graphicx}
\usepackage{booktabs}
\usepackage{textcomp}
\usepackage{xcolor}

\usepackage{subcaption}

\usepackage{bm}
\makeatletter
\def\ps@titlepage{%
  \def\@oddhead{}%
  \def\@evenhead{}%
  \def\@oddfoot{\hfill{\footerpagefont\thepage}\hfill}%
  \def\@evenfoot{\hfill{\footerpagefont\thepage}\hfill}%
}
\def\ps@headings{%
  \def\@oddhead{}%
  \def\@evenhead{}%
  \def\@oddfoot{\hfill{\footerpagefont\thepage}\hfill}%
  \def\@evenfoot{\hfill{\footerpagefont\thepage}\hfill}%
}
\def\headerlogo{}
\def\headerlogoall{}
\def\@pubid{}
\def\@history{}
\def\@doi{}
\def\@maketitle{%
     \setbox\maketitlebox\hbox\bgroup%
     \begin{minipage}{\titlewidth}%
     \raggedright\vspace*{1.2mm}%
     \ifaccesseditors\else\vspace*{-9.4mm}\par\fi%
     {\vss\color{accessblue}\titlefont\@title\vss}\vspace*{6mm}\par%
     \ifaccesseditors\vspace*{2mm}\else{\vss\authorfont\@author\vss}\vspace*{0mm}\par%
     \@tempcnta=0%
     \loop\ifnum\@tempcnta<\theaddrCtr%
       \advance\@tempcnta by 1%
       \par{\vss\addressfont\textsuperscript{\expandafter\csname addrInd\the\@tempcnta\endcsname}%
        \expandafter\csname addr\the\@tempcnta\endcsname\vss}%
     \repeat\vspace*{1mm}\par%
     {\vss\correspfont\@corresp\vss\vspace{2mm}}\par\fi%
     {\vss\tfootnotefont\@tfootnote\vss\vspace{7mm}}\par%
     \noindent\box99\vspace{6mm}\par%
     \ifaccesseditors\vspace*{8mm}\else%
     \noindent\box88\vspace{18mm}\par%
     \fi%
     \end{minipage}\par%
     \egroup%
     \fixSkip{\the\ht\maketitlebox}%
}
\AtBeginDocument{\pagestyle{headings}}
\makeatletter
\AtBeginDocument{\DeclareMathVersion{bold}
\SetSymbolFont{operators}{bold}{T1}{times}{b}{n}
\SetSymbolFont{NewLetters}{bold}{T1}{times}{b}{it}
\SetMathAlphabet{\mathrm}{bold}{T1}{times}{b}{n}
\SetMathAlphabet{\mathit}{bold}{T1}{times}{b}{it}
\SetMathAlphabet{\mathbf}{bold}{T1}{times}{b}{n}
\SetMathAlphabet{\mathtt}{bold}{OT1}{pcr}{b}{n}
\SetSymbolFont{symbols}{bold}{OMS}{cmsy}{b}{n}
\renewcommand\boldmath{\@nomath\boldmath\mathversion{bold}}}
\makeatother

\def\BibTeX{{\rm B\kern-.05em{\sc i\kern-.025em b}\kern-.08em
    T\kern-.1667em\lower.7ex\hbox{E}\kern-.125emX}}

\begin{document}

\title{Robust Lightweight Crack Classification for Real-Time UAV Bridge Inspection}
\author{\uppercase{Wei Li}\authorrefmark{1}, \uppercase{Haisheng Li}\authorrefmark{2},
\uppercase{Weijie Li}\authorrefmark{1}, \uppercase{Jiandong Wang}\authorrefmark{2},
\uppercase{Kaichen Ma}\authorrefmark{1}, and \uppercase{Luming Yang}\authorrefmark{2}}

\address[1]{Bay Area Super Bridge Maintenance Technology Center,
Guangdong Provincial Highway Construction Co., Ltd., Guangdong, China}
\address[2]{Guangdong AIHISUN Technology Co., Ltd., Guangdong, China}

\markboth
{Li \headeretal: Robust Lightweight Crack Classification for Real-Time UAV Bridge Inspection}
{Li \headeretal: Robust Lightweight Crack Classification for Real-Time UAV Bridge Inspection}

\corresp{E-mail: skylynfluke@gmail.com (Luming Yang)}

\begin{abstract}
With the widespread application of Unmanned Aerial Vehicles (UAVs) in bridge structural health monitoring, deep learning-based automatic crack detection has become a major research focus. However, practical UAV inspections still face four key challenges: weak crack features, degraded imaging conditions, severe class imbalance, and limited computational resources for practical UAV inspection workflows. To address these issues, this paper proposes a unified lightweight convolutional neural network framework composed of four synergistic components: a lightweight backbone network, a Convolutional Block Attention Module (CBAM) for channel and spatial enhancement, a directed robust augmentation strategy based on inspection-scene priors, and Focal Loss for hard-sample learning under class imbalance. Experiments on the SDNET2018 bridge deck dataset show that the proposed method achieves an inference speed of 825 FPS with only 11.21M parameters and 1.82G FLOPs. Compared with the baseline model, the complete framework improves the F1-score by 2.51\% and recall by 3.95\%. In addition, Grad-CAM visualizations indicate that the introduced attention module shifts the model's focus from scattered regions to precise tracking along crack trajectories. Overall, this study achieves a strong balance among accuracy, speed, and robustness, providing a practical solution for ground-station assisted real-time deployment in UAV bridge inspections. The source code is available at: https://github.com/skylynf/AttXNet . 
\end{abstract}


\titlepgskip=-21pt

\maketitle

\section{Introduction}

Bridges are indispensable components of modern transportation networks and play a fundamental role in sustaining regional connectivity, economic activity, and public safety. As bridge networks continue to age under long-term service conditions, routine condition assessment and structural health monitoring have become central tasks in bridge asset management. In practical engineering, local surface defects are often the most direct and accessible indicators of structural condition, among which concrete cracking is one of the most common and visually observable forms of deterioration. Crack information is of considerable importance for evaluating durability, identifying potential damage evolution, and supporting maintenance and rehabilitation decisions \cite{dong2021cvshm,luo2023bridgecv}.

For decades, bridge inspection has relied primarily on close visual examination supported by auxiliary access equipment. At the same time, non-destructive evaluation and structural health monitoring technologies have continuously expanded the technical toolkit available for bridge assessment. Within this broader development, computer vision has emerged as an attractive complement to conventional inspection because it enables remote, non-contact, and data-rich observation of structural surfaces and responses. Recent reviews have shown that vision-based approaches are now used not only for local defect identification, such as cracks, spalling, corrosion, and delamination, but also for broader monitoring tasks including displacement measurement, vibration analysis, and structural behavior characterization \cite{dong2021cvshm,luo2023bridgecv}. This evolution has laid an important foundation for image-based damage assessment.

The rapid development of Unmanned Aerial Vehicles (UAVs) has further accelerated this transformation. Early studies already demonstrated the feasibility of using UAV platforms for bridge-oriented sensing and visual servoing, showing that aerial systems could approach bridge surfaces and maintain appropriate viewpoints for inspection tasks \cite{metni2007uav}. Subsequent field investigations confirmed that UAVs can collect imagery suitable not only for qualitative observation but also for quantitative infrastructure evaluation, including crack visibility analysis, deformation-related measurement, and three-dimensional information acquisition \cite{ellenberg2015uav}. With continuing advances in flight control, onboard stabilization, camera resolution, and mission planning, UAVs have become an increasingly practical platform for bridge inspection, especially in areas that are difficult, hazardous, or inefficient to access by traditional means \cite{ham2016uavreview,feroz2021uavbridge,zhang2022fullyautomated}.

Beyond simple image capture, UAV-based bridge inspection has gradually expanded toward multi-source and digitalized data acquisition. Existing studies and reviews have reported the use of RGB imaging, infrared thermography, LiDAR, photogrammetry, and three-dimensional reconstruction for bridge condition assessment, enabling richer descriptions of decks, girders, piers, bearings, and other structural components \cite{feroz2021uavbridge,luo2023bridgecv,zhang2022fullyautomated}. In this context, UAV inspection is no longer viewed merely as a tool for visual documentation, but as an important carrier of high-quality sensing data for automated bridge evaluation.

In parallel with advances in data acquisition, automated crack recognition has evolved rapidly from handcrafted image processing methods to data-driven deep learning approaches. In the field of civil infrastructure inspection, convolutional neural networks and their derivatives have significantly improved the ability to learn crack-related texture, morphology, and contextual information directly from images. As a result, deep learning has been widely adopted for image-level crack classification, object-level defect detection, and pixel-level crack segmentation \cite{dong2021cvshm,dung2019fcn,islam2019encoderdecoder}. Compared with traditional thresholding-, filtering-, or edge-based techniques, deep networks provide a more unified learning framework for extracting discriminative crack features from complex concrete surfaces.

The maturation of public benchmark datasets has further promoted this research direction. A representative example is SDNET2018, which contains more than 56,000 annotated images of cracked and non-cracked concrete surfaces from bridge decks, walls, and pavements, and has become one of the most widely used datasets for evaluating concrete crack detection algorithms \cite{dorafshan2018sdnet}. With such datasets, the research community has been able to systematically compare different model architectures and training paradigms, thereby accelerating the development of practical crack inspection methods for real engineering scenarios.

Overall, the convergence of bridge structural health monitoring, computer vision, UAV-based sensing, and deep learning has created a strong technical basis for automated bridge crack inspection. From the perspective of bridge maintenance practice, UAV platforms provide efficient access to visual data; from the perspective of intelligent perception, deep neural networks provide increasingly powerful tools for extracting crack-related information from those data. This integration has made automated crack classification an important research direction for next-generation bridge inspection systems and an essential step toward more intelligent bridge management.

Although Convolutional Neural Network (CNN)-based visual detection methods have made substantial progress, practical UAV bridge inspection still faces four major challenges \cite{li2023automatic, zhou2025uav}:
\begin{enumerate}
    \item \textbf{Difficulty in Extracting Crack Features}: Early-stage cracks are usually thin and low-contrast, and are therefore easily confused with background scratches, water stains, and natural bridge-deck textures \cite{jiang2024advanced, yao2024cracknex}.
    \item \textbf{Poor Robustness to Imaging Environments}: Image quality often degrades due to UAV flight vibration, autofocus shifts \cite{lee2025optimizing}, and outdoor factors such as backlighting, shadows, and haze \cite{liu2020deep}.
    \item \textbf{Limited Edge-Computing Resources}: UAV onboard platforms have strict constraints on power and computation, making parameter-heavy and computationally intensive models difficult to deploy.
    \item \textbf{Severe Class Imbalance}: In large-scale inspection datasets, crack samples account for only a small proportion, causing models to be biased toward the majority non-crack class during training.
\end{enumerate}

Existing methods often fail to balance these constraints effectively: heavyweight models provide high accuracy but cannot support real-time onboard inference; lightweight models are efficient but frequently lack sufficient representation capacity for weak crack features; and generic augmentation pipelines are rarely tailored to inspection-specific degradation patterns \cite{hsieh2025development}.

\begin{figure}[htbp]
    \centering
    \includegraphics[width=0.9\linewidth]{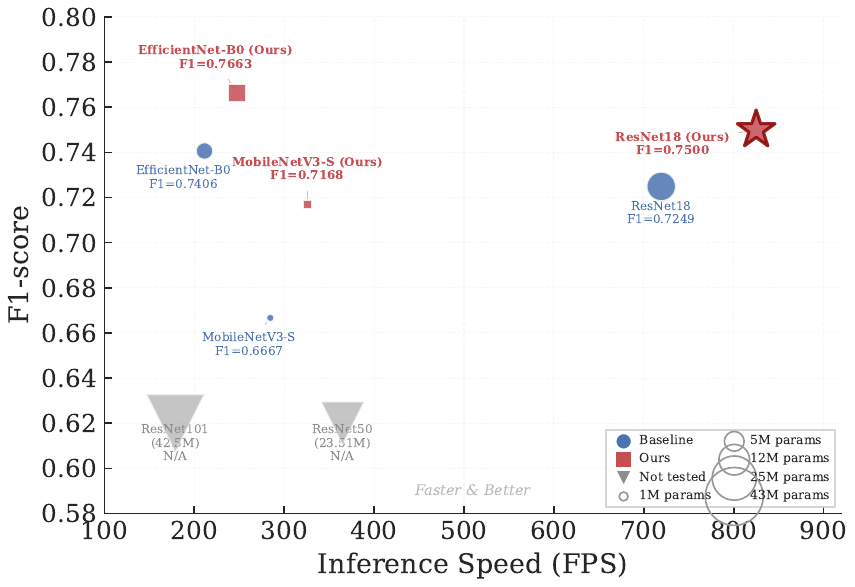}
    \caption{Three-dimensional comparison of FPS, test F1 score, and number of parameters (model size) among the baseline, our method, and untested references. Our framework achieves a favorable balance between efficiency and accuracy, improving overall model performance.}
    \label{fig:complexity}
\end{figure}

To address these practical limitations, this paper proposes a unified lightweight crack-classification framework for UAV-based bridge inspection, targeting an effective balance among accuracy, speed, and robustness. Rather than claiming a fundamentally new architecture, this work emphasizes deployable engineering integration. The main contributions are summarized as follows:
\begin{itemize}
    \item \textbf{Unified Deployment Framework for UAV Inspection:} We develop a highly integrated lightweight framework for UAV bridge inspection. By combining a lightweight backbone, attention, robust augmentation, and Focal Loss, the system satisfies real-time deployment constraints in a ground-station assisted workflow.
    \item \textbf{Directed Augmentation for Inspection Degradation:} We design a directed data-augmentation strategy guided by real inspection-scene priors to simulate typical degradation patterns, thereby reducing the distribution gap between training and deployment environments.
    \item \textbf{Systematic Empirical Study on Module Synergy:} We provide systematic empirical analysis of the functional roles and synergy among Robust Augmentation (RA), Focal Loss (FL), and CBAM. In particular, we identify the practical exclusivity between FL and class sampling, and show that a unified mechanism can handle both class imbalance and hard-sample mining.
    \item \textbf{Deployable Performance under Lightweight Constraints:} We demonstrate robust and practical performance as shown in Figure \ref{fig:complexity} under strict lightweight constraints. Extensive ablation studies, cross-backbone generalization, and Grad-CAM interpretability analysis verify both reliability and precise crack-trajectory tracking capability in real engineering scenarios.
\end{itemize}

\begin{figure*}[htbp]
    \centering
    \includegraphics[width=\linewidth]{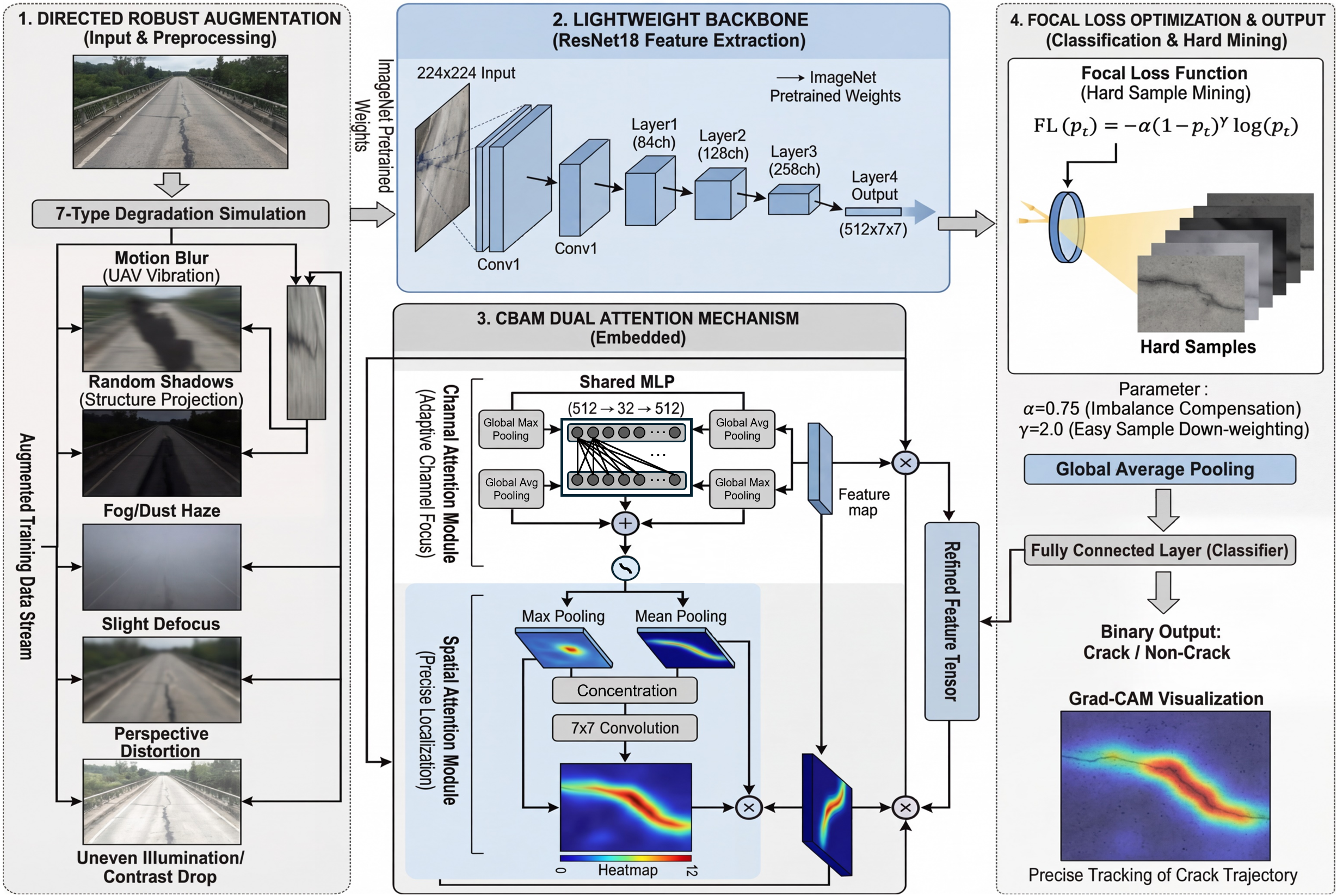}
    \caption{Overview of the proposed UAV-based bridge crack detection framework.
Raw bridge surface images are first processed through a data augmentation module
that simulates realistic environmental disturbances, including motion blur,
shadow, low illumination, fog, defocus blur, perspective distortion, and contrast degradation.
The augmented images are then fed into a ResNet18 backbone for hierarchical feature extraction.
A Convolutional Block Attention Module (CBAM) is integrated to refine feature representation
through sequential channel attention and spatial attention mechanisms.
The refined feature tensor is passed through global average pooling and a fully connected layer
to produce a binary classification result (crack / non-crack).
Focal Loss is employed to address class imbalance and emphasize hard samples.
Grad-CAM visualization is used to highlight crack localization and enhance model interpretability.}
    \label{fig:arch}
\end{figure*}


\subsection{UAV-based Infrastructure Inspection and Deep Learning}
The integration of Unmanned Aerial Vehicles (UAVs) and deep learning has revolutionized structural health monitoring. Traditional manual inspections are increasingly being replaced by UAV platforms equipped with high-definition cameras, which offer superior safety and efficiency. Recent studies have extensively explored Convolutional Neural Networks (CNNs) \cite{he2016deep} for crack detection in UAV imagery. For instance, Li et al. \cite{li2023automatic} demonstrated the efficacy of Faster R-CNN for automatic bridge crack detection using UAVs, highlighting the importance of stable imaging distances. More recent studies further broaden the technical landscape of UAV-enabled inspection and analysis. In addition to vision-based crack recognition, radar-assisted deep learning has been introduced for predictive beamforming and trajectory-aware UAV coordination in dynamic environments \cite{yin2025radar}, while graph neural networks and deep unfolding have been adopted for intelligent multi-UAV 3D trajectory and resource control \cite{yin2026intelligent}. In geotechnical and hazard-monitoring scenarios, UAV photogrammetry has also been integrated with advanced numerical modeling to characterize rockfall evolution and high-steep slope instability \cite{liu2026insights}. These latest studies show that current UAV research is moving toward tighter integration among perception, decision-making, and deployment-aware intelligence, which provides an important context for lightweight visual inspection systems.

\subsection{Lightweight Networks and Edge Deployment}
Although GPU-based algorithms can offer significant speed improvements \cite{10.1145/3638530.3654394}, a major bottleneck in real-time UAV inspection is the limited computational capacity and power supply of onboard edge devices \cite{pan2025cracklite, wang2025lightweight}. Consequently, deploying heavyweight models is often impractical. Researchers have increasingly focused on lightweight architectures to balance accuracy and inference speed \cite{yang2026multivariate,10.1145/3774906.3802786}. Zhou et al. \cite{zhou2025innovative} utilized a lightweight transfer-learning approach with EfficientNetB0 for real-time autonomous crack detection on drones. Similarly, Xiang et al. \cite{xiang2023gc} proposed a lightweight detector specifically optimized for UAV road crack detection, significantly reducing model parameters while maintaining performance. Recent 2024--2026 studies have further strengthened this direction. Yao et al. \cite{yao2024cracknex} proposed a few-shot low-light crack segmentation model for UAV inspection, showing the importance of efficiency under adverse illumination. Wang et al. \cite{wang2025lightweight} developed an improved MobileNetV3-based crack recognition model, and Pan et al. \cite{pan2025cracklite} reported a real-time lightweight network tailored to transportation-oriented crack detection. Beyond crack inspection, lightweight hybrid vision transformers with attention have also shown strong representational efficiency in visual classification tasks, indicating the growing value of transformer-assisted compact models for resource-constrained platforms \cite{wahid2026hybrid}. In this paper, we evaluate mainstream lightweight backbones, selecting ResNet18 \cite{he2016deep} as the optimal baseline due to its high inference speed and structural simplicity, which facilitates the independent evaluation of external attention modules.

\subsection{Attention Mechanisms for Weak Feature Extraction}
Early-stage cracks are often thin and exhibit low contrast, making them difficult to distinguish from natural concrete textures or scratches. To address this, attention mechanisms have been widely adopted to enhance the feature representation of neural networks \cite{11357122}. The Convolutional Block Attention Module (CBAM) \cite{woo2018cbam} is a prominent technique that sequentially infers attention maps along channel and spatial dimensions, effectively highlighting salient features while suppressing irrelevant background noise. Recent applications, such as the attention mechanisms explored by Ji et al. \cite{ji2025lightweight} for concrete crack detection, further validate the necessity of attention modules in complex structural environments. At the same time, transformer-based visual modeling has become an increasingly important direction because it captures long-range dependencies and global structural context more effectively than purely local convolutions. Hybrid lightweight transformer architectures with attention have recently achieved strong performance in visual feature extraction and classification \cite{wahid2026hybrid}, suggesting promising potential for crack inspection tasks where both local fine textures and broader contextual cues are important. Our framework integrates CBAM to precisely track crack trajectories without introducing significant computational overhead, while remaining compatible with future lightweight transformer extensions.

\subsection{Class Imbalance and Robust Training}
In real-world inspection datasets, such as the widely used SDNET2018 dataset \cite{maguire2018sdnet2018}, the ratio of crack to non-crack samples is extremely imbalanced. Standard cross-entropy loss often leads to models that are biased toward the majority background class. To counter this, Lin et al. \cite{lin2017focal} introduced Focal Loss, which dynamically scales the loss based on prediction confidence, forcing the model to focus on hard-to-classify examples. Furthermore, while generic data augmentation is common, it often fails to capture the specific degradations of UAV inspections. Recent studies have increasingly emphasized robust defect analysis under challenging acquisition conditions, including low-light crack inspection \cite{yao2024cracknex}, motion-blur enhancement in UAV concrete imagery \cite{liu2020deep}, and no-reference image-quality-guided optimization for mobile tunnel crack detection \cite{lee2025optimizing}. Robust geometric and structural analysis has also advanced rapidly in adjacent inspection domains: UAV photogrammetry has been combined with cone complementary-based 3D-DDA to analyze rockfall evolution \cite{liu2026insights}, and 3D laser scanning has been integrated with discontinuity identification and DDA simulation to improve reliability in hazard characterization \cite{liu2026three}. For binary classification tasks involving imbalanced datasets, there are also methods based on recombination, which underscores the importance of environmental augmentation \cite{10.1093/bib/bbaf609}. Together, these developments highlight that modern defect analysis is evolving toward robust, multi-condition, and deployment-aware systems. Our directed augmentation strategy follows the same practical motivation by simulating typical UAV imaging degradations and working synergistically with Focal Loss to achieve robust performance in real-world deployments \cite{ma2021real, seibold2017model}.

\section{Methodology}

The overall pipeline of the proposed lightweight deep learning framework for UAV bridge crack recognition is as follows (as shown in Figure \ref{fig:arch}): the input image first passes through a lightweight backbone network to extract basic feature maps, followed by adaptive weighting via an attention module. Finally, the features are passed through global pooling and a fully connected classifier to output a binary classification result (crack/non-crack).

\subsection{Lightweight Backbone}
To meet the real-time requirements of UAV onboard deployment, this paper evaluates three mainstream lightweight CNNs as feature extraction backbones: ResNet18 \cite{he2016deep}, MobileNetV3-Small \cite{howard2019searching}, and EfficientNet-B0 \cite{tan2019efficientnet}. 

In subsequent core ablation experiments, \textbf{ResNet18} is selected as the baseline backbone. The primary rationale is that ResNet18 possesses an extremely high inference speed (up to 928 FPS for the pure backbone) and its native architecture does not contain built-in attention mechanisms (unlike MobileNetV3 and EfficientNet-B0, which include Squeeze-and-Excitation modules). This characteristic allows the gains from the external attention module (CBAM) \cite{woo2018cbam} to be observed independently and clearly, avoiding functional redundancy. All backbone networks are initialized with ImageNet pre-trained weights to accelerate convergence.

\subsection{Attention Enhancement Module (CBAM)}
To enhance the model's ability to capture weak crack features, a Convolutional Block Attention Module (CBAM) is inserted after the final feature map output by the backbone network (e.g., the layer4 output of ResNet18, which has 512 channels and a \(7 \times 7\) spatial resolution). CBAM consists of two sequential sub-modules:
\begin{enumerate}
    \item \textbf{Channel Attention}: Global average pooling and global max pooling are applied separately along the spatial dimensions of the feature map. The two resulting vectors pass through a shared two-layer Multi-Layer Perceptron (MLP, with dimension changes \(512 \rightarrow 32 \rightarrow 512\)). The outputs are summed and activated by a Sigmoid function to obtain a channel weight vector, which is element-wise multiplied with the original feature map. This enables the network to adaptively focus on "which feature channels are most sensitive to cracks."
    \item \textbf{Spatial Attention}: Mean and max aggregations are performed along the channel dimension. The concatenated result is reduced to a single channel via a \(7 \times 7\) convolution, followed by Sigmoid activation to generate a spatial weight map. This map is element-wise multiplied with the channel attention output, allowing the network to precisely locate "where cracks exist in the spatial dimensions."
\end{enumerate}
It is worth emphasizing that introducing the CBAM module only adds 0.03M parameters (a 0.27\% increase), and the FLOPs remain almost unchanged, making its impact on inference speed negligible.

\subsection{Focal Loss for Hard Sample Optimization}
In practical inspection data, the ratio of crack to non-crack samples is extremely imbalanced (approximately \(1:5.7\) in our dataset). Standard Cross-Entropy (CE) loss treats all samples equally, causing the model to bias toward the majority class. Therefore, this paper introduces Focal Loss \cite{lin2017focal}:
\[
\text{FL}(p_t) = -\alpha_t (1 - p_t)^\gamma \log(p_t)
\]
Here, the class balancing factor \(\alpha = 0.75\) is assigned to the minority class (cracks) to directly compensate for class imbalance. The focusing parameter \(\gamma = 2.0\) ensures that the \((1-p_t)^\gamma\) term significantly down-weights the loss of high-confidence, easily classified samples. This mechanism forces the model to automatically focus on hard-to-distinguish, blurry cracks and difficult samples corrupted by degradation during training.

\subsection{Inspection-Scene Robust Training Strategy}

In this study, domain shift primarily refers to the covariate shift in visual appearance between images of reference concrete surfaces and actual images captured during UAV inspections. Typical sources of shift include lighting variations, low contrast, motion blur, defocus, haze-like degradation, shadows, perspective distortion, and image noise. The robust enhancement strategy proposed in this paper explicitly randomizes these factors during the training process, thereby improving the model's robustness to actual UAV imaging conditions.

Unlike generic data augmentation methods, this paper designs a directed simulation strategy targeting 7 typical imaging degradations common in UAV bridge inspections, aiming to narrow the distribution gap between training data and real-world images:
\begin{enumerate}
    \item \textbf{Uneven Illumination/Contrast Drop} (Probability 0.5): Simulates backlit or shaded shooting.
    \item \textbf{Motion Blur} (3-7px, Probability 0.3): Simulates UAV flight vibrations.
    \item \textbf{Mild Defocus} (3-5px, Probability 0.2): Simulates autofocus shifts.
    \item \textbf{Fog/Dust} (Probability 0.2): Simulates dusty or foggy environments.
    \item \textbf{Random Shadows} (Probability 0.25): Simulates projections from the bridge structure itself.
    \item \textbf{Perspective Distortion} (3-8\%, Probability 0.2): Simulates non-orthogonal viewing angles.
    \item \textbf{Color Jitter/Low Light} (Probability 0.3/0.2): Simulates evening or cloudy lighting conditions.
\end{enumerate}

\subsection{Real-world WORKFLOW AND DEPLOYMENT SETTING}

\begin{figure*}[htbp]
    \centering
    \includegraphics[width=0.8\linewidth]{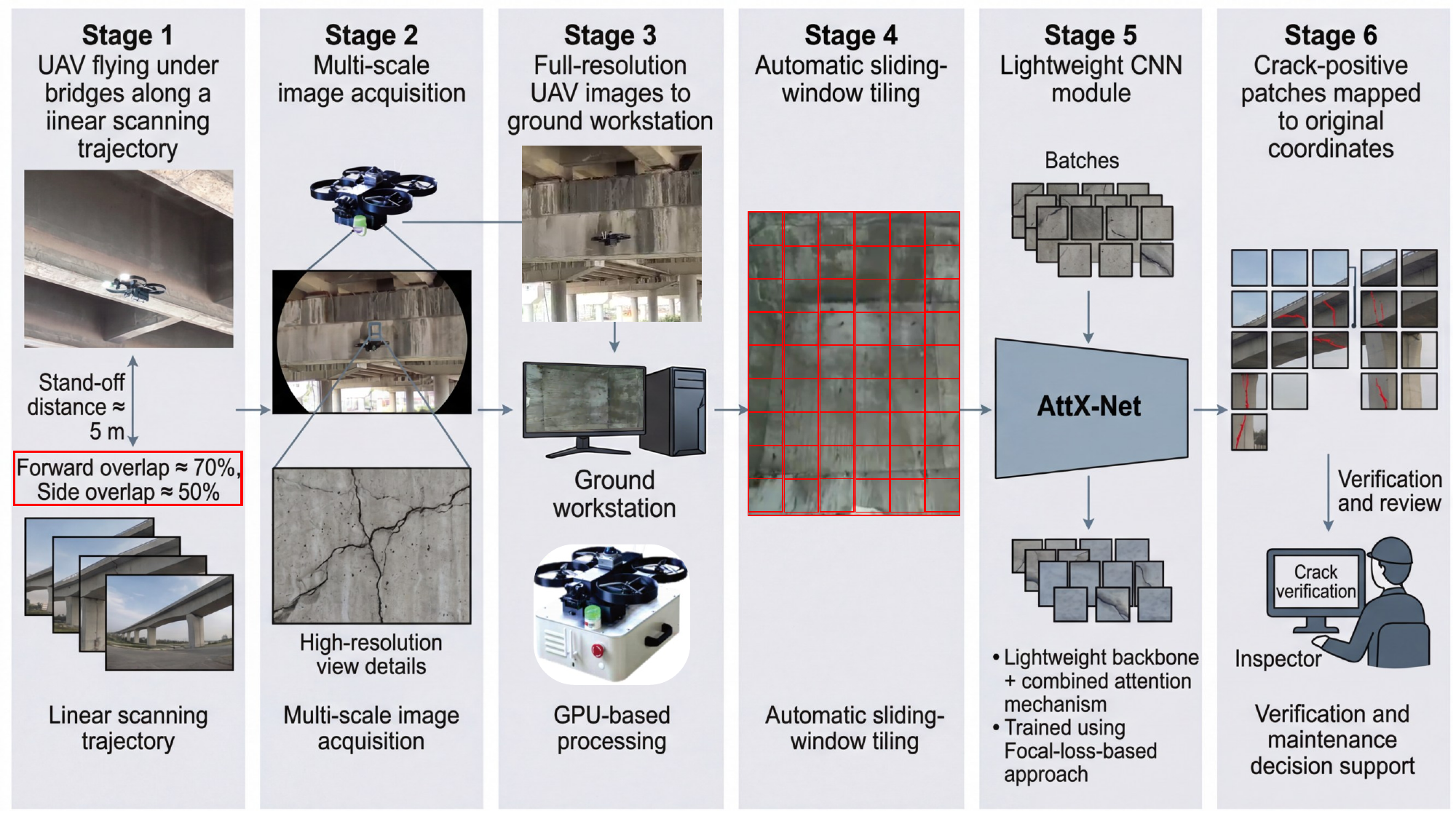}
    \caption{Practical ground-station-assisted UAV bridge inspection workflow. The UAV continuously acquires high-resolution bridge images along a pre-planned route. The photos show our actual deployment scenario.
}
    \label{fig:uav_workflow}
\end{figure*}

It should be noted that the proposed network is designed as a patch-level crack classification module rather than a standalone object detection, segmentation, or crack quantification system. Given a cropped bridge-surface image patch, the model outputs a binary prediction, i.e., crack or non-crack. To bridge the gap between patch-level classification and full-scene UAV bridge inspection, this module is embedded into a ground-station-assisted UAV inspection workflow, as illustrated in Figure \ref{fig:uav_workflow}.

During field inspection, the UAV follows a pre-planned continuous scanning route along the bridge surface or underside. The UAV does not hover at each suspicious crack location for immediate human confirmation. Instead, it continuously acquires high-resolution images with sufficient overlap, thereby improving data acquisition efficiency and reducing field operation complexity. In the practical inspection experiment, the UAV maintained an approximate stand-off distance of 5 m from the inspected bridge surface. The forward overlap and side overlap were approximately 70\% and 50\%, respectively. A short exposure time greater than \(1/1000\,\mathrm{s}\) was adopted to reduce motion blur caused by UAV vibration and platform movement. Under this imaging configuration, the estimated ground sampling distance was approximately \(0.1\sim0.2\,\mathrm{mm/pixel}\), which is sufficient for observing fine surface cracks in local image patches.

The image acquisition strategy follows a photogrammetry-level multi-scale receptive-field design. Wide-angle images provide global structural context and can support panoramic stitching or three-dimensional reconstruction, while close-range or telephoto images provide local high-resolution details for crack recognition. After image acquisition, full-resolution UAV images are transmitted to a ground workstation for near-real-time or near-offline batch processing. Therefore, the reported high-speed inference in this study corresponds to a ground-station deployment setting rather than fully validated onboard UAV edge inference.

In the ground-station processing stage, full-resolution UAV images are automatically divided into fixed-size patches using a sliding-window tiling strategy. Thus, candidate regions are generated automatically rather than manually selected. The proposed lightweight classifier is then applied to these patches in batch mode. Patches predicted as crack-positive are retained as candidate crack regions, while non-crack patches are discarded or assigned a lower inspection priority.

Finally, each positive patch is mapped back to the original full-resolution image according to its cropping coordinates. If panoramic stitching or photogrammetric reconstruction is available, the same coordinate relationship can also be used to project candidate crack regions onto the reconstructed bridge surface model. Although the network itself performs binary patch classification, this coordinate-mapping step enables full-image-level crack-candidate localization. Inspectors can then review the automatically screened candidate regions for verification and maintenance decision support. Therefore, the proposed method should be understood as a lightweight crack-screening and localization-support module within a practical UAV bridge inspection pipeline, rather than as a complete crack detection or quantification system.

\subsection{Overall Training Algorithm}

The overall training procedure is summarized in Algorithm \ref{alg:training}.

\begin{algorithm}[htbp]
\caption{Bridge Crack Classification Training}
\label{alg:training}
\begin{algorithmic}[1]
\Require Training images \(\mathcal{X}\), labels \(\mathcal{Y}\)
\Ensure Trained model \(\mathcal{M}\)
\State Load ResNet18 pretrained on ImageNet as backbone \(\mathcal{B}\)
\State Attach CBAM module after \(\mathcal{B}\).layer4
\State Attach classifier head: GlobalAvgPool \(\rightarrow\) Dropout(0.2) \(\rightarrow\) FC(512 \(\rightarrow\) 2)
\State Define loss: FocalLoss(\(\alpha=0.75, \gamma=2.0\))
\State Define optimizer: AdamW(lr=\(1e-3\), weight\_decay=\(1e-4\))
\State Define scheduler: CosineAnnealing + 3-epoch LinearWarmup
\For{\(epoch = 1\) \textbf{to} \(40\)}
    \For{each batch \((x, y)\) in train\_loader}
        \State \(\tilde{x} \gets \text{RobustAugmentation}(x)\) \Comment{7-type degradation}
        \State \(f \gets \mathcal{B}.\text{forward\_features}(\tilde{x})\) \Comment{\((B, 512, 7, 7)\)}
        \State \(f \gets \text{CBAM}(f)\) \Comment{Attention-weighted features}
        \State \(\hat{y} \gets \text{Classifier}(\text{GlobalAvgPool}(f))\) 
        \State \(\mathcal{L} \gets \text{FocalLoss}(\hat{y}, y)\)
        \State Backpropagate and update weights
    \EndFor
    \State Evaluate on validation set, save best model by F1-score
\EndFor
\State Test on held-out test set
\end{algorithmic}
\end{algorithm}

\section{Experiments and Results Analysis}

\subsection{Experimental Setup}
Experiments are conducted on the deck category (D) of the public SDNET2018 dataset \cite{maguire2018sdnet2018}. The dataset contains 2,025 crack images and 11,595 non-crack images (a ratio of \(1:5.7\)). The data is split into 70\% training, 15\% validation, and 15\% testing sets using stratified sampling. Input images are uniformly resized to \(224 \times 224\) resolution. Training utilizes the AdamW optimizer \cite{loshchilov2017decoupled} (learning rate 1e-3, weight decay 1e-4) with a cosine annealing learning rate scheduler and a 3-epoch warmup \cite{loshchilov2016sgdr}. The total number of training epochs is 40, with a batch size of 64. The hardware platform is equipped with 3 \(\times\) NVIDIA RTX 4090 (24GB) GPUs.

\subsection{Lightweight Backbone Comparison}
\begin{figure*}[htbp]
    \centering
    \includegraphics[width=0.8\linewidth]{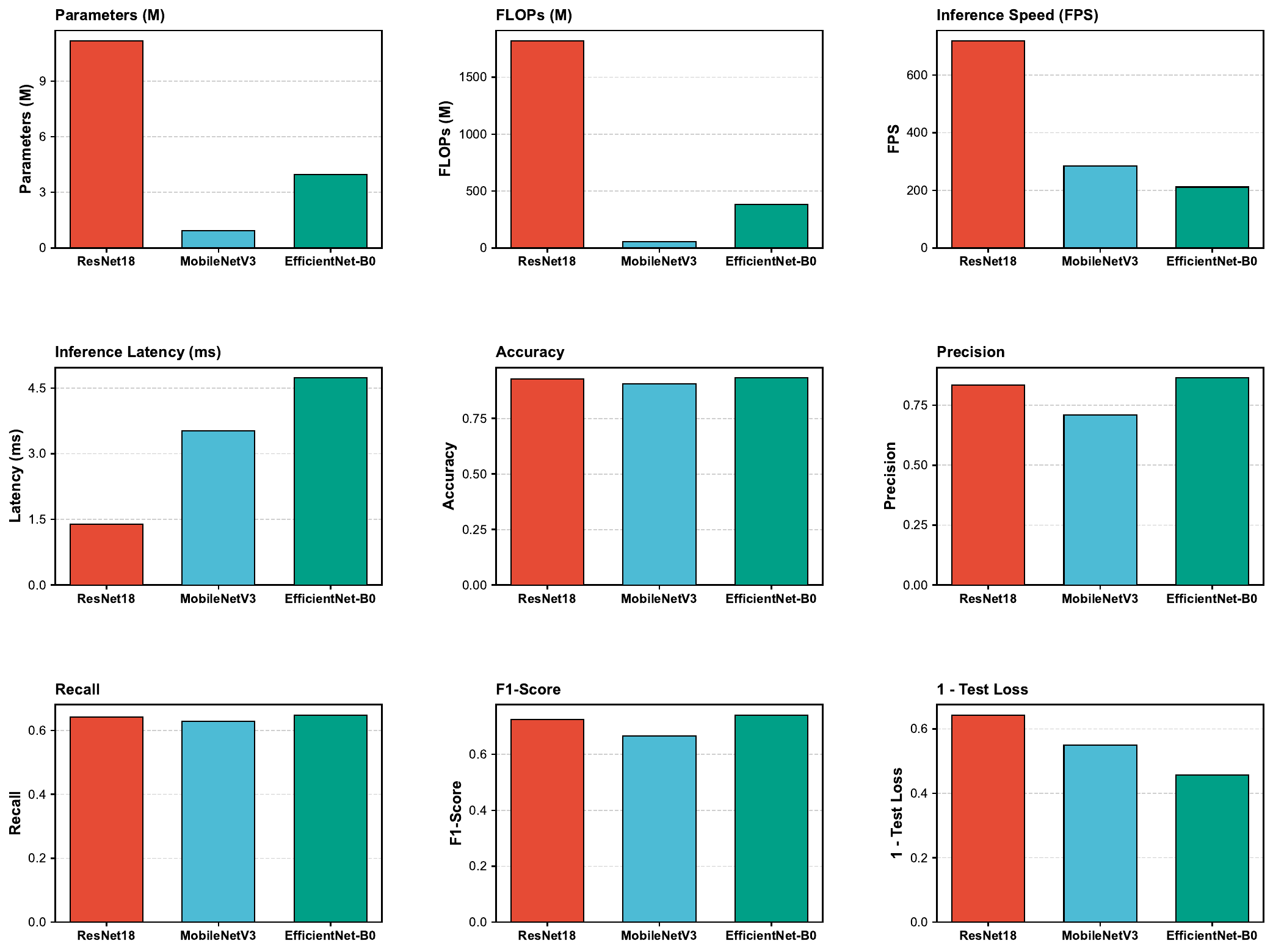}
    \caption{Comparison of resource consumption, inference efficiency, and classification performance of three backbone networks (bar chart matrix).
ResNet18, MobileNetV3, and EfficientNet-B0 are compared in terms of Parameters, FLOPs, inference speed (FPS), inference latency (Latency), and five classification metrics including Accuracy, Precision, Recall, F1-score, and $1 - \text{Test Loss}$. 
The bar chart matrix provides a comprehensive evaluation of the backbone networks from three perspectives: resource consumption, computational efficiency, and predictive performance.
}
    \label{fig:33}
\end{figure*}

As shown in Figure \ref{fig:33} and \ref{fig:radar}, all three lightweight backbones achieve over 90\% accuracy, indicating that transfer learning with pre-trained weights is highly effective for bridge crack classification. EfficientNet-B0 yields the best overall performance (F1=0.7406) but has the slowest inference speed (211.4 FPS). Considering the strict real-time requirements of UAV onboard deployment, ResNet18 provides the optimal speed-accuracy trade-off, achieving an inference speed of 719.4 FPS (3.4 times faster than EfficientNet) and an F1-score of 0.7249. Therefore, it is selected as the baseline backbone for subsequent ablation studies. Although MobileNetV3 has extremely few parameters, its shallow feature dimensions lead to suboptimal performance on this dataset.


\subsection{Core Ablation and Imbalance-Handling Study}

To verify the effectiveness of each component and to further examine the influence of different imbalance-handling strategies, we conducted a step-by-step ablation study based on the ResNet18 baseline, as shown in Table \ref{tab:ablation}. The evaluated variants include single-component settings, alternative class-imbalance strategies, and the progressively combined versions of the proposed framework.

Compared with the baseline cross-entropy model, using Robust Augmentation (RA) alone increases the Recall from 0.6414 to 0.7467, achieving the highest Recall among all variants. This indicates that the degradation simulation used in RA effectively improves the model's sensitivity to crack regions under non-ideal inspection conditions. However, this improvement is accompanied by a clear decrease in Precision, from 0.8333 to 0.7276, suggesting that RA also introduces more false positives.

Focal Loss (FL) alone does not improve the overall F1-score over the baseline, obtaining an F1-score of 0.7138. This suggests that imbalance-aware loss reweighting by itself is insufficient when the model has not yet learned sufficiently robust crack representations. In contrast, when FL is combined with RA, the F1-score increases to 0.7548, which is the best F1-score in the ablation study and represents a 2.99 percentage-point improvement over the baseline. This result shows that RA and FL are complementary: RA enhances crack sensitivity, while FL helps suppress easy or dominant samples and improves the Precision--Recall trade-off.

We also compare FL with two commonly used imbalance-handling alternatives: inverse-frequency weighted cross-entropy and weighted sampling. Weighted cross-entropy improves the F1-score to 0.7324, while weighted sampling further improves it to 0.7446. Both methods outperform the baseline, confirming the importance of addressing class imbalance. Nevertheless, they remain inferior to the RA + FL configuration, demonstrating that the proposed combination provides a stronger balance between robustness and class-imbalance mitigation.

Finally, adding the CBAM attention module to RA + FL further increases Accuracy to 0.9325 and Precision to 0.8347, both of which are the highest values among all variants. Although the F1-score slightly decreases from 0.7548 to 0.7500 due to a reduction in Recall, the resulting model provides a more conservative and precise prediction behavior. In practical bridge inspection scenarios, such high Precision is valuable because it reduces false alarms and lowers the manual verification burden. Therefore, while RA + FL achieves the highest F1-score, the final model with CBAM is selected as a deployment-oriented trade-off among Precision, interpretability, and robustness.

The role of each component can be summarized as follows:
\begin{enumerate}
    \item \textbf{Robust Augmentation (RA)} improves crack sensitivity and substantially increases Recall by simulating realistic inspection degradations such as illumination variation and image quality deterioration.
    
    \item \textbf{Focal Loss (FL)} is most effective when combined with RA. It helps balance the Precision--Recall trade-off by reducing the dominance of easy majority-class samples, leading to the best F1-score in the ablation study.
    
    \item \textbf{CBAM Attention} improves the reliability and compactness of crack-related feature localization. It achieves the highest Precision and Accuracy, making the final model more suitable for deployment scenarios where false alarms should be minimized.
\end{enumerate}

\begin{figure}[htbp]
    \centering
    \includegraphics[width=0.7\linewidth]{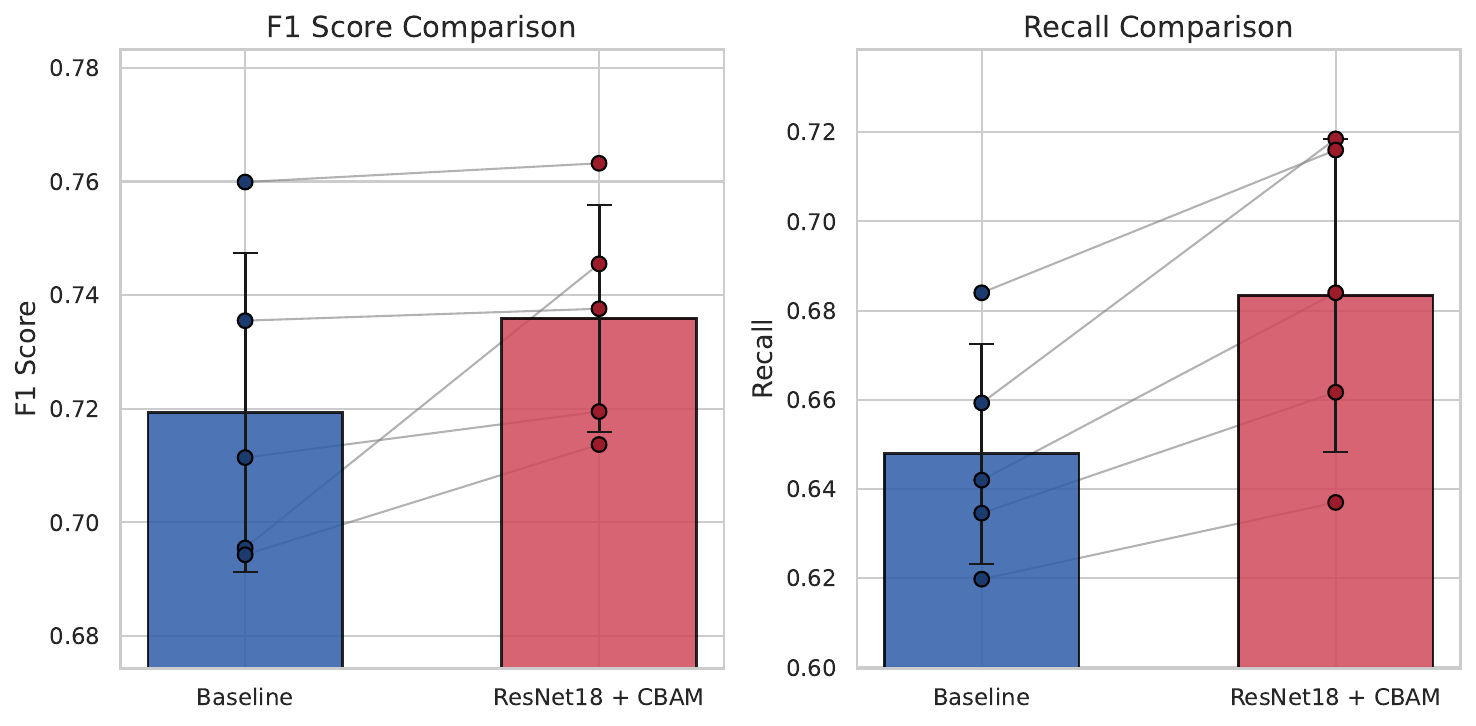}
    \caption{Comparison of baseline ResNet18 and CBAM-enhanced ResNet18 under five-fold cross-validation. Bars represent mean performance (F1-score and Recall), dots indicate individual fold results, and error bars denote standard deviation. Statistical significance annotations are included.}
    \label{fig:five_fold_cv_statistics}
\end{figure}

\begin{table*}[htbp]
\centering
\caption{Ablation Study Based on ResNet18}
\label{tab:ablation}
\begin{tabular}{lccccccc}
\toprule
\textbf{Configuration} & \textbf{Accuracy} & \textbf{Precision} & $\Delta$Precision & \textbf{Recall} & $\Delta$Recall & \textbf{F1} & $\Delta$F1 \\
\midrule
Baseline (CE) & 0.9276 & 0.8333 & — & 0.6414 & — & 0.7249 & — \\

+ FL only & 0.9227 & 0.7944 & -4.67\% & 0.6480 & +1.03\% & 0.7138 & -1.11\% \\

+ RA only & 0.9207 & 0.7276 & -12.69\% & \textbf{0.7467} & +16.42\% & 0.7370 & +1.21\% \\

+ Weighted CE (inv. freq.) & 0.9256 & 0.7879 & -5.45\% & 0.6842 & +6.67\% & 0.7324 & +0.75\% \\

+ Weighted Sampler & 0.9305 & 0.8214 & -1.43\% & 0.6809 & +6.16\% & 0.7446 & +1.97\% \\

+ RA + FL & 0.9310 & 0.8007 & -3.91\% & 0.7138 & +11.29\% & \textbf{0.7548} & +2.99\% \\

+ RA + FL + CBAM & \textbf{0.9325} & \textbf{0.8347} & +0.17\% & 0.6809 & +6.16\% & 0.7500 & +2.51\% \\
\bottomrule
\end{tabular}
\end{table*}
\subsection{Attention Module Comparison}
\begin{figure}[htbp]
    \centering
    \includegraphics[width=\linewidth]{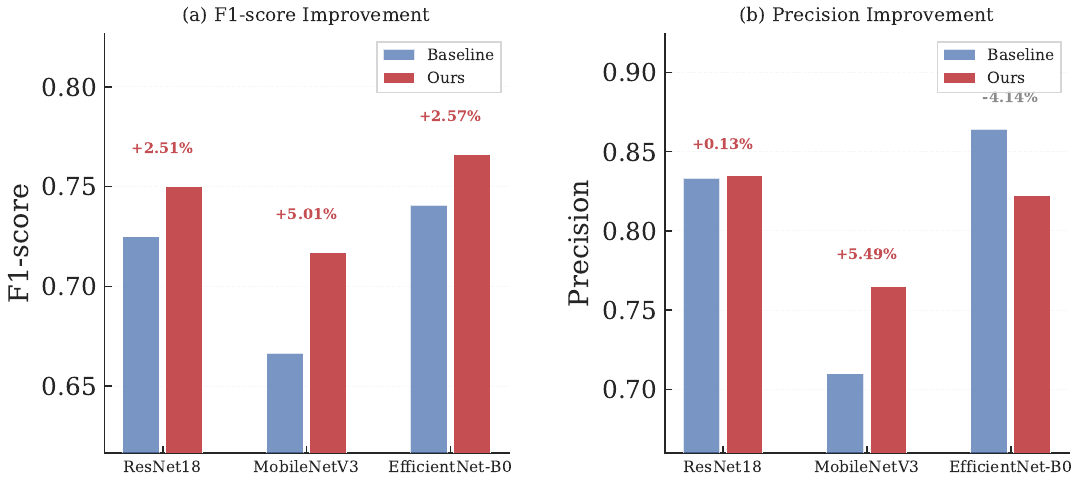}
    \caption{Comparison of F1-score and precision between the baseline and our method on three backbone networks. The annotated values indicate the percentage improvement (\(\Delta\%\)) over the baseline. Recall results are reported separately in the table below.}
    \label{fig:f1precision}
\end{figure}

Table \ref{tab:attention} compares the effects of different attention mechanisms. CBAM achieves the best results in both Accuracy (0.9325) and Precision (0.8347), demonstrating that its dual channel-spatial attention mechanism can more accurately locate discriminative crack features. In contrast, Coordinate Attention (CA) approaches the no-attention baseline in Recall but lacks sufficient Precision. Considering the importance of high Precision in actual deployment, CBAM is the superior choice. See comparisons in Figure \ref{fig:f1precision}

To further evaluate the robustness of the proposed method, we conducted five-fold cross-validation and performed paired statistical tests between the baseline ResNet18 model and the CBAM-enhanced ResNet18 model, as shown in Figure \ref{fig:five_fold_cv_statistics}.

Across all folds, the CBAM-enhanced ResNet18 consistently outperformed the baseline model. Specifically, the average F1-score improved from 0.7193 to 0.7359, while the recall increased more substantially from 0.6479 to 0.6835.

Paired statistical analysis demonstrated a significant improvement in recall, as confirmed by a paired t-test (\( p = 0.0076 \)) and a large effect size (Cohen's \( d = 2.23 \)). The F1-score also showed a clear improvement trend, supported by the Wilcoxon signed-rank test (\( p = 0.0625 \)). Since there are only five folds, \( p = 0.0625 \) is the smallest p possible.

These findings indicate that the CBAM module primarily enhances the sensitivity to crack-positive samples without negatively affecting overall classification stability.

\begin{table}[htbp]
\centering
\caption{Comparison of Attention Modules}
\label{tab:attention}
\begin{tabular}{lcccc}
\toprule
\textbf{Attention} & \textbf{Accuracy} & \textbf{Precision} & \textbf{Recall} & \textbf{F1} \\
\midrule
None & 0.9310 & 0.8007 & \textbf{0.7138} & \textbf{0.7548} \\
CBAM & \textbf{0.9325} & \textbf{0.8347} & 0.6809 & 0.7500 \\
CA & 0.9242 & 0.7578 & 0.7204 & 0.7386 \\
\bottomrule
\end{tabular}
\end{table}

\subsection{Cross-Backbone Generalization}
Table \ref{tab:generalization} proves the universality of our framework. The complete method demonstrates consistent performance improvements across all three lightweight backbones, with F1-score increases ranging from 2.51\% to 5.01\% and Recall increases from 3.95\% to 6.91\%. Notably, even on architectures with built-in SE channel attention like MobileNetV3 and EfficientNet-B0, our method still yields significant improvements. This is attributed to the spatial attention dimension provided by CBAM—while the SE module only focuses on "which channels are important," CBAM simultaneously focuses on "where it is important spatially," forming an excellent complement.

\begin{table}[htbp]
\centering
\caption{Cross-Backbone Generalization Validation}
\label{tab:generalization}
\begin{tabular}{lcccc}
\toprule
\textbf{Backbone} & \textbf{Base F1} & \textbf{Ours F1} & $\Delta$F1 & $\Delta$Recall \\
\midrule
ResNet18 & 0.7249 & 0.7500 & +2.51\% & +3.95\% \\
MobileNetV3 & 0.6667 & 0.7168 & \textbf{+5.01\%} & +4.61\% \\
EfficientNet-B0 & 0.7406 & \textbf{0.7663} & +2.57\% & \textbf{+6.91\%} \\
\bottomrule
\end{tabular}
\end{table}

\subsection{Parameter Analysis}

\begin{figure}[htbp]
    \centering
    \includegraphics[width=0.8\linewidth]{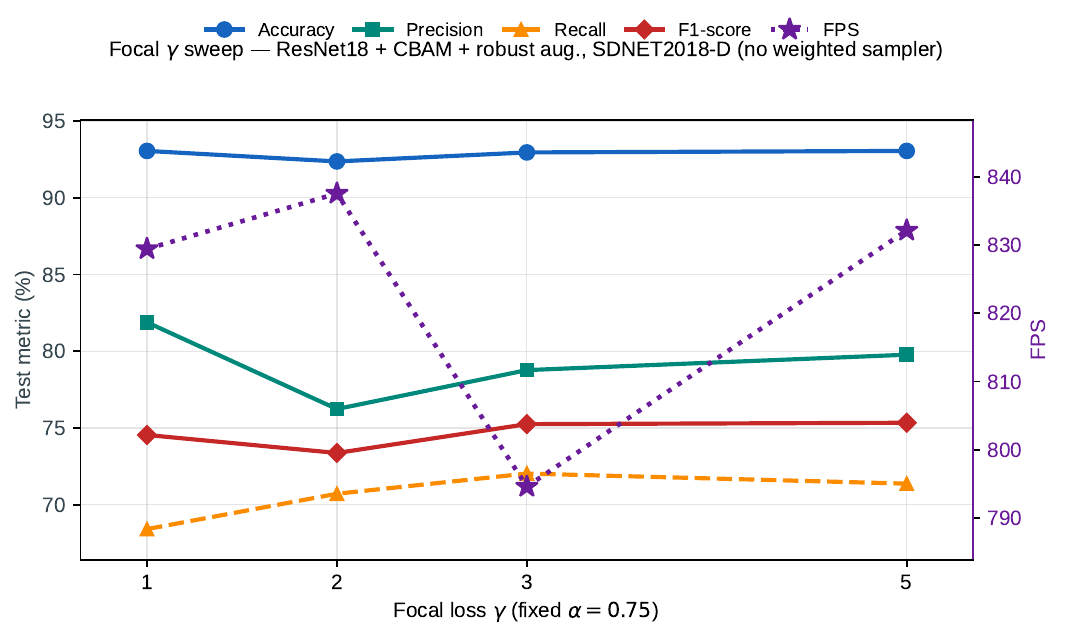}
    \caption{Parameter comparison on Focal loss $\gamma$ under ResNet18 + CBAM + RA.}
    \label{fig:param}
\end{figure}

The results (Figure \ref{fig:param}) show that the model performance remains stable within a reasonable parameter range. In particular, $\gamma=3.0$ and $\alpha=0.75$ provide a favorable balance between recall improvement and precision preservation under the observed class imbalance ratio. Excessively small $\gamma$ weakens hard-sample mining, whereas excessively large $\gamma$ may overemphasize difficult or noisy samples and reduce precision. Similarly, excessively aggressive augmentation may degrade visual discriminability, while moderate UAV-scene augmentation improves robustness.

\subsection{Computational Complexity Analysis}
As shown in Table \ref{tab:complexity}, the proposed ResNet18+CBAM method requires only 11.21M parameters and 1.82G FLOPs, which is less than half the parameters of ResNet50 and one-quarter of ResNet101, while achieving an extremely high inference speed of 825 FPS (latency of only 1.21ms). The CBAM module introduces merely 0.03M additional parameters, providing significant performance enhancements with almost no impact on inference efficiency.

\begin{figure}[t]
    \centering
    
    \begin{subfigure}{0.9\linewidth}
        \centering
        \includegraphics[width=\linewidth]{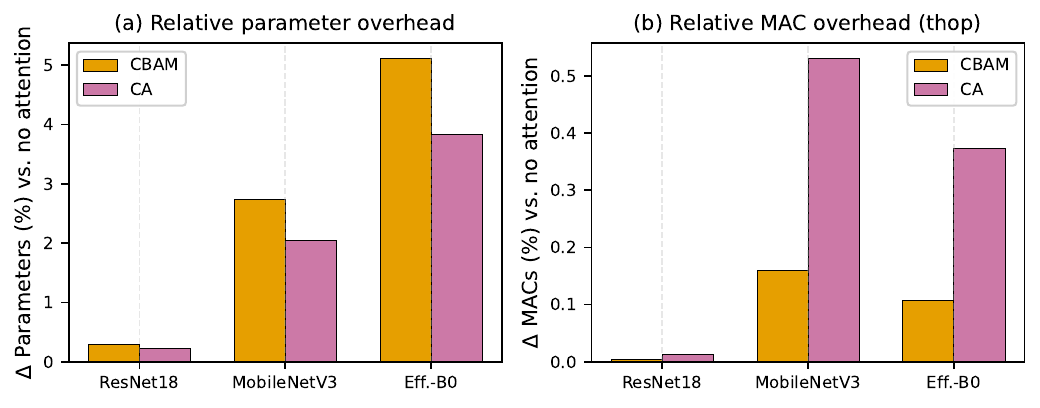}
    \end{subfigure}
    
    \vspace{0.5cm}
    
    \begin{subfigure}{0.9\linewidth}
        \centering
        \includegraphics[width=\linewidth]{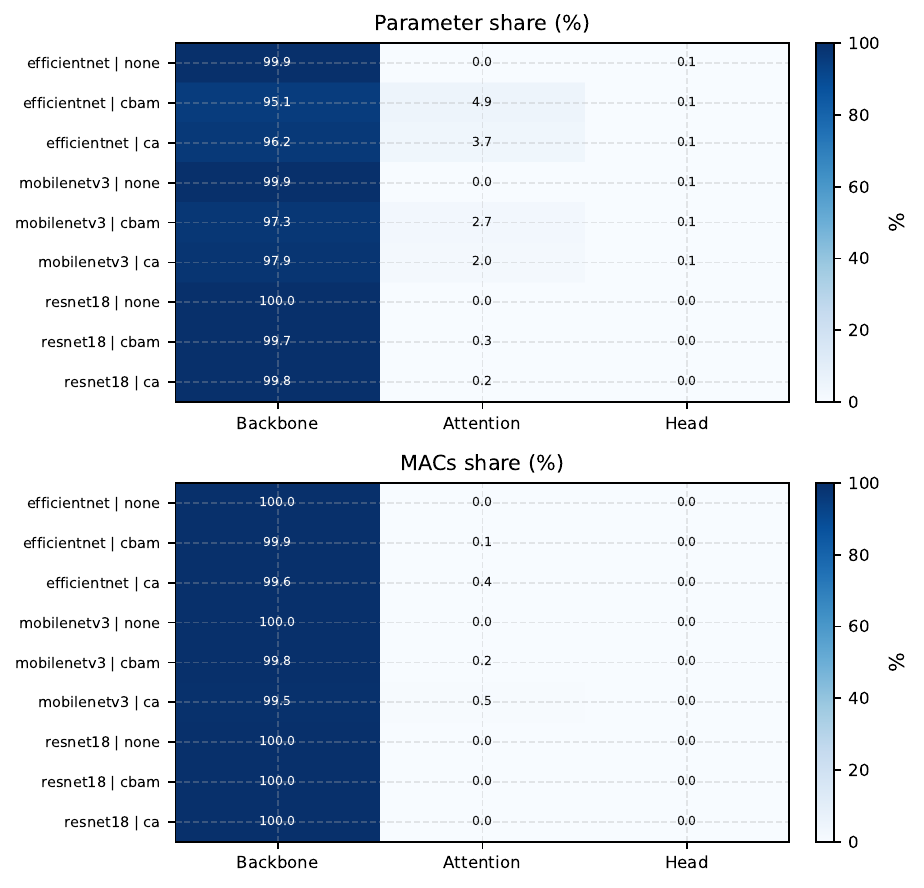}
    \end{subfigure}
    
    \caption{Comparison of attention overhead across different backbones. 
    The top figure shows the parameter and MAC increase introduced by the attention module (boxplot), 
    the bottom heatmap illustrates the relative parameter and MAC shares of backbone, attention module, 
    and classifier head.}
    
    \label{fig:attention_overhead_comparison}
\end{figure}

With a fixed input resolution of \(224 \times 224\), we decompose each network into a backbone, an external attention module, and a linear classifier head. The results shown in Figure \ref{fig:attention_overhead_comparison} indicate that the backbone dominates in terms of both parameter count and MACs. The classifier head contributes only a parameter share of approximately \(10^{-4} \sim 10^{-3}\), while its contribution to MACs is negligible. Compared to their corresponding attention-free baselines, CBAM introduces only a limited increase in parameters, and the relative increase in MACs across all tested backbones remains below approximately 0.6\%. These results demonstrate that the proposed attention-enhanced lightweight framework improves crack recognition performance while maintaining practical inference efficiency.

\begin{table}[htbp]
\centering
\caption{Computational Complexity Comparison}
\label{tab:complexity}
\begin{tabular}{lcccc}
\toprule
\textbf{Model} & \textbf{Params} & \textbf{FLOPs} & \textbf{FPS} & \textbf{Latency} \\
\midrule
\textbf{ResNet18+CBAM (Ours)} & \textbf{11.21M} & \textbf{1.82G} & \textbf{825} & \textbf{1.21ms} \\
MobileNetV3+CBAM & 958K & 55.0M & 326 & 3.06ms \\
EfficientNet+CBAM & 4.17M & 385.0M & 247 & 4.05ms \\
ResNet50 (Heavyweight) & 23.51M & 4.13G & 365 & 2.74ms \\
ResNet101 (Heavyweight) & 42.50M & 7.86G & 179 & 5.58ms \\
\bottomrule
\end{tabular}
\end{table}

\begin{figure}[htbp]
    \centering
    \includegraphics[width=\linewidth]{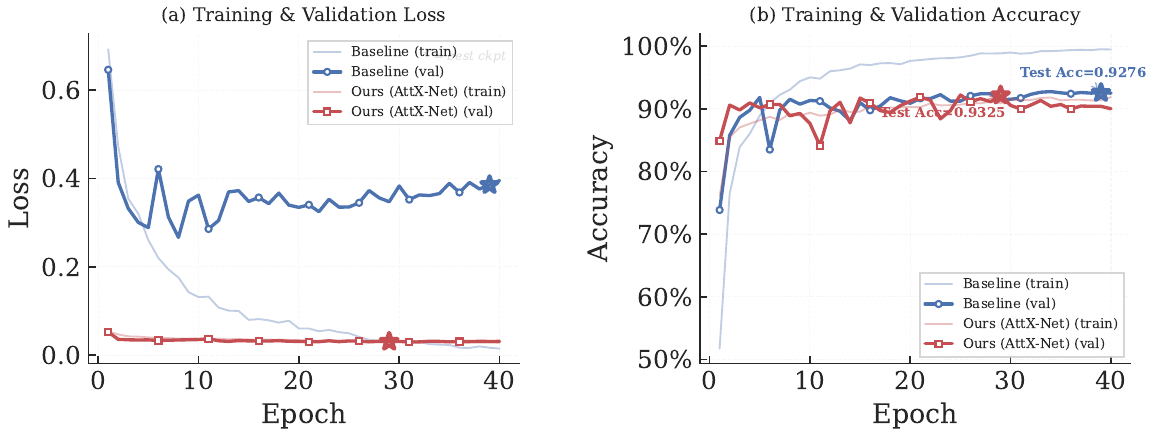}
    \caption{Training and validation loss/accuracy curves of the baseline and our method on ResNet18. The star marker denotes the automatically selected best epoch (final checkpoint). The test accuracy at this epoch is also annotated.}
    \label{fig:training_curves}
\end{figure}

\begin{figure*}[htbp]
    \centering
    \includegraphics[width=\linewidth]{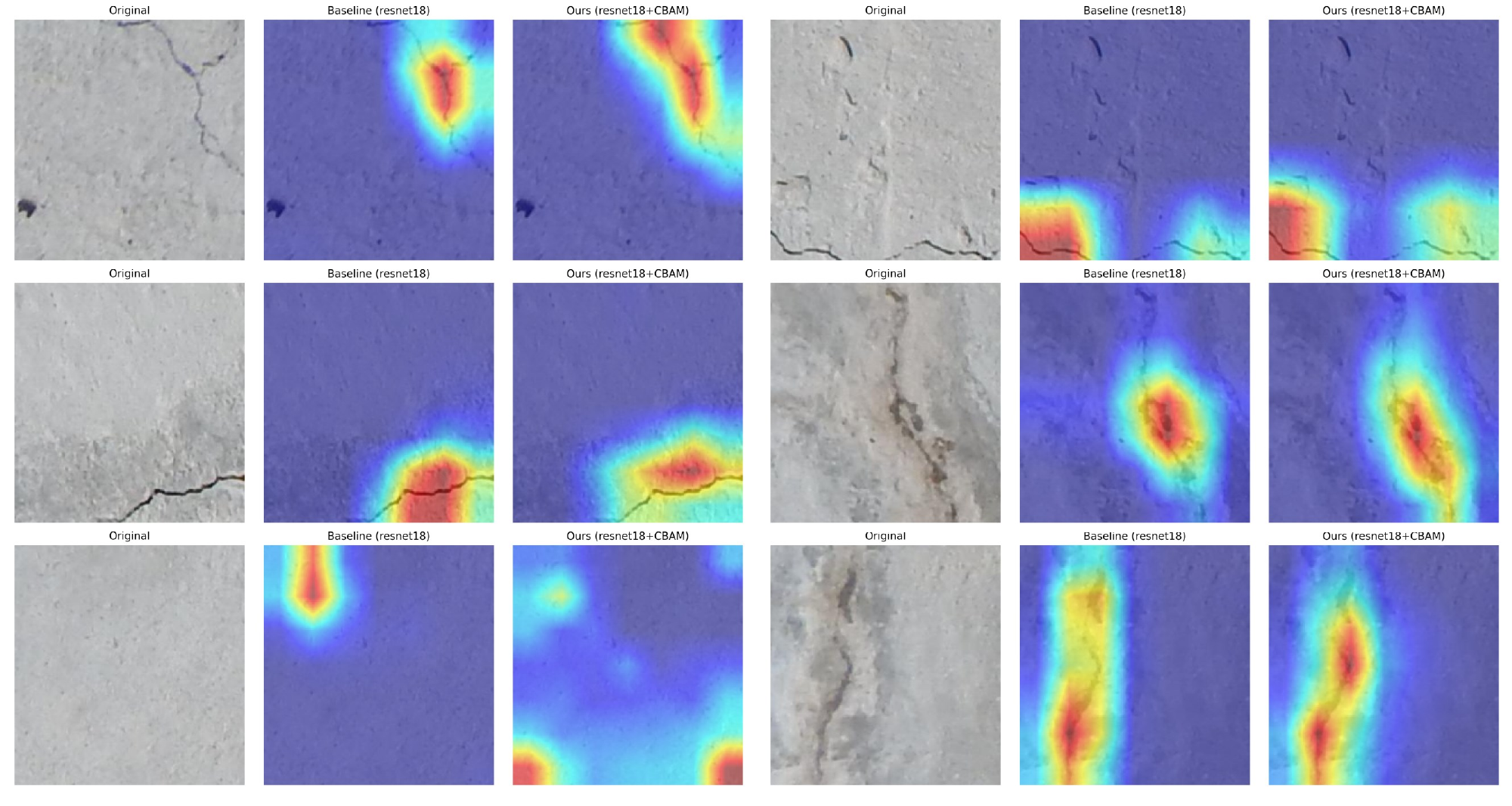}
    \caption{Grad-CAM visualizations comparing the baseline model and the proposed AttX-Net. The attention maps demonstrate precise tracking along the crack trajectory and effective background suppression.}
    \label{fig:gradcam}
\end{figure*}

\begin{figure}[htbp]
    \centering
    \includegraphics[width=\linewidth]{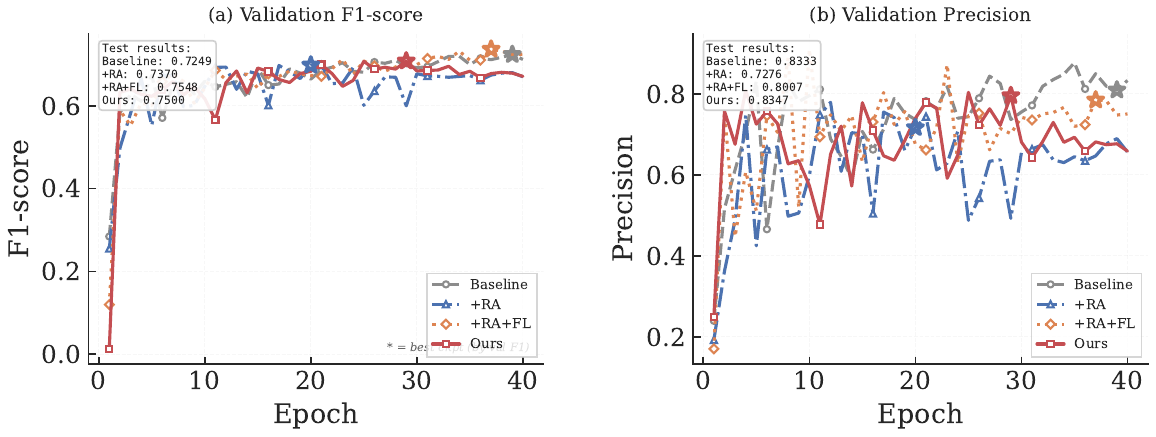}
    \caption{Validation F1-score and precision across epochs for different ablation stages. The star marker indicates the automatically selected best epoch used for final evaluation. The inset summarizes the test results at that epoch.}
    \label{fig:metric_curves}
\end{figure}

\subsection{Training Dynamics Analysis}
Observations of the training curves (Figure \ref{fig:training_curves} and \ref{fig:metric_curves}) indicate that both the baseline model and the proposed method converge stably within 40 epochs without signs of overfitting. Notably, the training-validation accuracy gap of the complete method (approx. 3\%) is smaller than that of the baseline model (approx. 5\%). This suggests that the inspection-scene robust augmentation strategy not only increases data diversity but also acts as an implicit regularizer, effectively enhancing the model's generalization performance.

After incorporating our method particularly with the inclusion of RA, the training and validation curves exhibit noticeably larger oscillations. This behavior is expected and reasonable. RA introduces stronger stochastic perturbations to the input data, effectively increasing the diversity and difficulty of training samples at each epoch. As a result, the optimization process is exposed to continuously varying data distributions, which leads to higher variance in intermediate performance metrics. However, this increased fluctuation reflects enhanced regularization rather than instability. By preventing the model from overfitting to specific patterns and encouraging it to learn more robust and generalized representations, RA ultimately contributes to improved generalization performance on the test set.

\subsection{Grad-CAM Interpretability Analysis}

To further investigate the working mechanism of the attention module, Grad-CAM visualizations (Figure \ref{fig:gradcam}) were extracted from the final feature map of the network. Compared with the baseline model, the network optimized with CBAM demonstrates substantially improved attention distribution in several aspects:

\begin{enumerate}
    \item \textbf{Directional Tracking Capability}: The baseline model tends to concentrate attention on isolated local points, whereas the proposed method expands attention along the crack trajectory, effectively covering the entire crack path and preserving structural continuity.
    
    \item \textbf{Multi-Target Sensitivity}: While the baseline model typically responds to only a single local region, the proposed method produces strong activations across multiple potential crack regions, thereby enhancing the detection of weak and subtle cracks.
    
    \item \textbf{Precise Localization and Background Suppression}: The attention maps generated by the proposed method exhibit clearer and more compact boundaries, effectively suppressing irrelevant background activations and improving spatial focus.
\end{enumerate}

These qualitative observations provide a clear explanation for the substantial improvement in Precision (from 0.8007 to 0.8347) observed in the quantitative evaluation. More accurate attention localization directly reduces false positive predictions, thereby enhancing overall detection reliability.

\subsection{Limitations}

Although Grad-CAM visualizations suggest improved attention localization and background suppression, our interpretability analysis remains primarily qualitative. The absence of pixel-level annotations prevents us from conducting quantitative faithfulness or localization-based interpretability evaluations.

Although stratified tile-level splitting is commonly used for benchmarking evaluation on SDNET2018, it may not completely eliminate potential spatial correlations between visually similar regions. For future UAV-based datasets that include explicit bridge location metadata, structure-level or inspection-region-level splitting should be adopted to more rigorously evaluate generalization across physically distinct locations.

\section{Conclusions}

To address key practical challenges in UAV-based bridge inspection, this paper presents a lightweight attention-augmented convolutional neural network framework (AttX-Net). By integrating a lightweight backbone, CBAM dual attention, Focal Loss, and a robust training strategy guided by real-world degradation priors, the proposed framework achieves a favorable trade-off between accuracy and computational cost. Experimental results show that, with only 11.21M parameters and an inference speed of 825 FPS, the method improves crack-recognition F1-score by 2.51\% and recall by 3.95\% over the baseline, satisfying real-time processing requirements in a ground-station assisted UAV inspection workflow. Grad-CAM visualizations further confirm that the model can track crack trajectories more precisely and suppress irrelevant background responses.

Future work will proceed in three directions: (1) extending the lightweight framework to pixel-level crack segmentation, (2) introducing multi-task learning to jointly model crack-type classification and severity assessment, and (3) validating long-term closed-loop performance through deployment on physical UAV platforms.


\section*{Code and data availability}

The implementation details, training configuration, data-splitting scripts, and Grad-CAM visualization code will be released at https://github.com/skylynf/AttXNet upon publication. The SDNET2018 dataset used in this study is publicly available. Additional real UAV inspection images used for workflow verification are not publicly released due to engineering-site restrictions, but representative examples and processing procedures are provided in the revised manuscript.

\bibliographystyle{IEEEtran}
\bibliography{attx}

\begin{thebibliography}{10}
\providecommand{\url}[1]{#1}
\csname url@samestyle\endcsname
\providecommand{\newblock}{\relax}
\providecommand{\bibinfo}[2]{#2}
\providecommand{\BIBentrySTDinterwordspacing}{\spaceskip=0pt\relax}
\providecommand{\BIBentryALTinterwordstretchfactor}{4}
\providecommand{\BIBentryALTinterwordspacing}{\spaceskip=\fontdimen2\font plus
\BIBentryALTinterwordstretchfactor\fontdimen3\font minus \fontdimen4\font\relax}
\providecommand{\BIBforeignlanguage}[2]{{%
\expandafter\ifx\csname l@#1\endcsname\relax
\typeout{** WARNING: IEEEtran.bst: No hyphenation pattern has been}%
\typeout{** loaded for the language `#1'. Using the pattern for}%
\typeout{** the default language instead.}%
\else
\language=\csname l@#1\endcsname
\fi
#2}}
\providecommand{\BIBdecl}{\relax}
\BIBdecl

\bibitem{dong2021cvshm}
C.-Z. Dong and F.~N. Catbas, ``A review of computer vision-based structural health monitoring at local and global levels,'' \emph{Structural Health Monitoring}, vol.~20, no.~2, pp. 692--743, 2021.

\bibitem{luo2023bridgecv}
K.~Luo, X.~Kong, J.~Zhang, J.~Hu, J.~Li, and H.~Tang, ``Computer vision-based bridge inspection and monitoring: A review,'' \emph{Sensors}, vol.~23, no.~18, p. 7863, 2023.

\bibitem{metni2007uav}
N.~Metni and T.~Hamel, ``A {UAV} for bridge inspection: Visual servoing control law with orientation limits,'' \emph{Automation in Construction}, vol.~17, no.~1, pp. 3--10, 2007.

\bibitem{ellenberg2015uav}
A.~Ellenberg, L.~Branco, A.~Krick, I.~Bartoli, and A.~Kontsos, ``Use of unmanned aerial vehicle for quantitative infrastructure evaluation,'' \emph{Journal of Infrastructure Systems}, vol.~21, no.~3, p. 04014054, 2015.

\bibitem{ham2016uavreview}
Y.~Ham, K.~K. Han, J.~J. Lin, and M.~Golparvar-Fard, ``Visual monitoring of civil infrastructure systems via camera-equipped unmanned aerial vehicles ({UAV}s): a review of related works,'' \emph{Visualization in Engineering}, vol.~4, p.~1, 2016.

\bibitem{feroz2021uavbridge}
S.~Feroz and S.~Abu~Dabous, ``{UAV}-based remote sensing applications for bridge condition assessment,'' \emph{Remote Sensing}, vol.~13, no.~9, p. 1809, 2021.

\bibitem{zhang2022fullyautomated}
C.~Zhang, Y.~Zou, F.~Wang, E.~del Rey~Castillo, J.~Dimyadi, and L.~Chen, ``Towards fully automated unmanned aerial vehicle-enabled bridge inspection: Where are we at?'' \emph{Construction and Building Materials}, vol. 347, p. 128543, 2022.

\bibitem{dung2019fcn}
C.~V. Dung and L.~D. Anh, ``Autonomous concrete crack detection using deep fully convolutional neural network,'' \emph{Automation in Construction}, vol.~99, pp. 52--58, 2019.

\bibitem{islam2019encoderdecoder}
M.~M.~M. Islam and J.-M. Kim, ``Vision-based autonomous crack detection of concrete structures using a fully convolutional encoder--decoder network,'' \emph{Sensors}, vol.~19, no.~19, p. 4251, 2019.

\bibitem{dorafshan2018sdnet}
S.~Dorafshan, R.~J. Thomas, and M.~Maguire, ``{SDNET2018}: An annotated image dataset for non-contact concrete crack detection using deep convolutional neural networks,'' \emph{Data in Brief}, vol.~21, pp. 1664--1668, 2018.

\bibitem{li2023automatic}
R.~Li, J.~Yu, F.~Li, R.~Yang, Y.~Wang, and Z.~Peng, ``Automatic bridge crack detection using unmanned aerial vehicle and faster r-cnn,'' \emph{Construction and Building Materials}, vol. 362, p. 129659, 2023.

\bibitem{zhou2025uav}
L.~Zhou, Y.~Jiang, H.~Jia, L.~Zhang, F.~Xu, Y.~Tian, Z.~Ma, X.~Liu, S.~Guo, Y.~Wu \emph{et~al.}, ``Uav vision-based crack quantification and visualization of bridges: system design and engineering application,'' \emph{Structural Health Monitoring}, vol.~24, no.~2, pp. 1083--1100, 2025.

\bibitem{jiang2024advanced}
T.~Jiang, L.~Liu, C.~Hu, L.~Li, and J.~Zheng, ``An advanced method for surface damage detection of concrete structures in low-light environments based on image enhancement and object detection networks,'' \emph{Advances in Bridge Engineering}, vol.~5, no.~1, p.~33, 2024.

\bibitem{yao2024cracknex}
Z.~Yao, J.~Xu, S.~Hou, and M.~C. Chuah, ``Cracknex: a few-shot low-light crack segmentation model based on retinex theory for uav inspections,'' in \emph{2024 IEEE International Conference on Robotics and Automation (ICRA)}.\hskip 1em plus 0.5em minus 0.4em\relax IEEE, 2024, pp. 11\,155--11\,162.

\bibitem{lee2025optimizing}
C.~Lee, D.~Kim, and D.~Kim, ``Optimizing deep learning-based crack detection using no-reference image quality assessment in a mobile tunnel scanning system,'' \emph{Sensors}, vol.~25, no.~17, p. 5437, 2025.

\bibitem{liu2020deep}
Y.~Liu, J.~K. Yeoh, and D.~K. Chua, ``Deep learning--based enhancement of motion blurred uav concrete crack images,'' \emph{Journal of computing in civil engineering}, vol.~34, no.~5, p. 04020028, 2020.

\bibitem{hsieh2025development}
H.-Y. Hsieh, K.-Y. Liu, and S.~Kang, ``Development of an automated surface crack detection and bim-integrated management system for concrete bridges,'' \emph{Journal of Civil Engineering and Management}, vol.~31, no.~7, pp. 710--728, 2025.

\bibitem{he2016deep}
K.~He, X.~Zhang, S.~Ren, and J.~Sun, ``Deep residual learning for image recognition,'' in \emph{Proceedings of the IEEE conference on computer vision and pattern recognition}, 2016, pp. 770--778.

\bibitem{yin2025radar}
R.~Yin, J.~Peng, Y.~Cai, C.~Wu, B.~Champagne, and N.~Al-Dhahir, ``Radar-assisted predictive beamforming for uav-aided networks: a deep-learning solution,'' \emph{IEEE Transactions on Vehicular Technology}, 2025.

\bibitem{yin2026intelligent}
------, ``Intelligent 3d trajectory and resource control for multi-uav 6g networks via gnn and deep unfolding,'' \emph{IEEE Transactions on Communications}, 2026.

\bibitem{liu2026insights}
G.~Liu, J.~Liu, H.~Fan, S.~He, W.~Bo, C.~Yang, and J.~Miao, ``Insights into evolution of rockfalls on a high-steep slope using uav photogrammetry and cone complementary-based 3d-dda,'' \emph{Canadian Geotechnical Journal}, no.~ja, 2026.

\bibitem{10.1145/3638530.3654394}
\BIBentryALTinterwordspacing
L.~Yang, T.~Jiang, and R.~Cheng, ``Tensorized ant colony optimization for gpu acceleration,'' in \emph{Proceedings of the Genetic and Evolutionary Computation Conference Companion}, ser. GECCO '24 Companion.\hskip 1em plus 0.5em minus 0.4em\relax New York, NY, USA: Association for Computing Machinery, 2024, p. 755–758. [Online]. Available: \url{https://doi.org/10.1145/3638530.3654394}
\BIBentrySTDinterwordspacing

\bibitem{pan2025cracklite}
R.~Pan and Y.~Zhang, ``Cracklite-net: A sustainable transportation-oriented real-time lightweight network for adaptive road crack detection,'' \emph{Sustainability}, vol.~17, no.~24, p. 10973, 2025.

\bibitem{wang2025lightweight}
R.~Wang, R.~Chen, H.~Yan, and X.~Guo, ``Lightweight concrete crack recognition model based on improved mobilenetv3,'' \emph{Scientific Reports}, vol.~15, no.~1, p. 15704, 2025.

\bibitem{yang2026multivariate}
L.~Yang, H.~Liu, S.~Li, and A.~Yilmaz, ``Multivariate gaussian representation learning for medical action evaluation,'' in \emph{Proceedings of the AAAI Conference on Artificial Intelligence}, vol.~40, no.~46, 2026, pp. 39\,513--39\,521.

\bibitem{10.1145/3774906.3802786}
\BIBentryALTinterwordspacing
Y.~Liu, H.~Jiang, H.~Liu, R.~Huang, and X.~Ouyang, ``Movid: View-invariant 3d human pose estimation via motion-view disentanglement,'' in \emph{Proceedings of the 2026 ACM/IEEE International Conference on Embedded Artificial Intelligence and Sensing Systems}, ser. SenSys '26.\hskip 1em plus 0.5em minus 0.4em\relax New York, NY, USA: Association for Computing Machinery, 2026, p. 1194–1207. [Online]. Available: \url{https://doi.org/10.1145/3774906.3802786}
\BIBentrySTDinterwordspacing

\bibitem{zhou2025innovative}
C.~Zhou, M.~Dai, F.~Wang, Y.~Dong, X.~Chen, and C.~He, ``An innovative uav and deep learning-based framework for automatic bridge crack detection and measurement,'' \emph{The Journal of Supercomputing}, vol.~81, no.~15, p. 1410, 2025.

\bibitem{xiang2023gc}
X.~Xiang, H.~Hu, Y.~Ding, Y.~Zheng, and S.~Wu, ``Gc-yolov5s: a lightweight detector for uav road crack detection,'' \emph{Applied Sciences}, vol.~13, no.~19, p. 11030, 2023.

\bibitem{wahid2026hybrid}
A.~Wahid, H.~U. Khan, A.~Naz, and F.~K. Alarfaj, ``Hybrid lightweight vision transformers with attention mechanism for feature extraction and classification of product designs,'' \emph{Plos one}, vol.~21, no.~3, p. e0343510, 2026.

\bibitem{11357122}
J.~Hong, S.~Li, M.~Jian, and L.~Yang, ``Bidirectional time-frequency pyramid network for enhanced robust eeg classification,'' in \emph{2025 IEEE International Conference on Bioinformatics and Biomedicine (BIBM)}, 2025, pp. 2225--2232.

\bibitem{woo2018cbam}
S.~Woo, J.~Park, J.-Y. Lee, and I.~S. Kweon, ``Cbam: Convolutional block attention module,'' in \emph{Proceedings of the European conference on computer vision (ECCV)}, 2018, pp. 3--19.

\bibitem{ji2025lightweight}
H.~Ji, Z.~Zeng, and X.~Dong, ``Lightweight concrete crack detection for urban intelligent management and maintenance,'' in \emph{Proceedings of the Institution of Civil Engineers-Transport}.\hskip 1em plus 0.5em minus 0.4em\relax Emerald Publishing Limited, 2025.

\bibitem{maguire2018sdnet2018}
M.~Maguire, S.~Dorafshan, and R.~J. Thomas, ``Sdnet2018: A concrete crack image dataset for machine learning applications,'' 2018.

\bibitem{lin2017focal}
T.-Y. Lin, P.~Goyal, R.~Girshick, K.~He, and P.~Doll{\'a}r, ``Focal loss for dense object detection,'' in \emph{Proceedings of the IEEE international conference on computer vision}, 2017, pp. 2980--2988.

\bibitem{liu2026three}
G.~Liu, J.~Kang, T.~Ye, S.~Wang, W.~Bo, D.~Duoji, and Y.~Tian, ``Three-dimensional (3d) laser scanning--based identification of rock mass discontinuities for rockfall modeling using 3d discontinuous deformation analysis,'' \emph{International Journal of Rock Mechanics and Mining Sciences}, vol. 202, p. 106484, 2026.

\bibitem{10.1093/bib/bbaf609}
\BIBentryALTinterwordspacing
L.~Yang, H.~Liu, A.~Calanche, S.~M. Gokcek, V.~Singh, N.~Sansoterra, M.~Akkaya, B.~Akkaya, and A.~Yilmaz, ``Disrupting explicit encoding paradigms: property-interactive transformers decode t-cell receptor specificity beyond dataset biases,'' \emph{Briefings in Bioinformatics}, vol.~26, no.~6, p. bbaf609, 11 2025. [Online]. Available: \url{https://doi.org/10.1093/bib/bbaf609}
\BIBentrySTDinterwordspacing

\bibitem{ma2021real}
Y.~Ma, Q.~Li, L.~Chu, Y.~Zhou, and C.~Xu, ``Real-time detection and spatial localization of insulators for uav inspection based on binocular stereo vision,'' \emph{Remote Sensing}, vol.~13, no.~2, p. 230, 2021.

\bibitem{seibold2017model}
C.~Seibold, A.~Hilsmann, and P.~Eisert, ``Model-based motion blur estimation for the improvement of motion tracking,'' \emph{Computer Vision and Image Understanding}, vol. 160, pp. 45--56, 2017.

\bibitem{howard2019searching}
A.~Howard, M.~Sandler, G.~Chu, L.-C. Chen, B.~Chen, M.~Tan, W.~Wang, Y.~Zhu, R.~Pang, V.~Vasudevan \emph{et~al.}, ``Searching for mobilenetv3,'' in \emph{Proceedings of the IEEE/CVF international conference on computer vision}, 2019, pp. 1314--1324.

\bibitem{tan2019efficientnet}
M.~Tan and Q.~Le, ``Efficientnet: Rethinking model scaling for convolutional neural networks,'' in \emph{International conference on machine learning}.\hskip 1em plus 0.5em minus 0.4em\relax PMLR, 2019, pp. 6105--6114.

\bibitem{loshchilov2017decoupled}
I.~Loshchilov and F.~Hutter, ``Decoupled weight decay regularization,'' \emph{arXiv preprint arXiv:1711.05101}, 2017.

\bibitem{loshchilov2016sgdr}
------, ``Sgdr: Stochastic gradient descent with warm restarts,'' \emph{arXiv preprint arXiv:1608.03983}, 2016.

\end{thebibliography}

\EOD

\end{document}